%% file: main.tex
\documentclass[10pt]{article}

\PassOptionsToPackage{svgnames}{xcolor}
\usepackage{style}

\usepackage[numbers,sort&compress]{natbib}
\usepackage{graphicx}
\usepackage[font=small,labelfont=bf]{caption}
\usepackage{subcaption}
\usepackage{booktabs}
\usepackage{colortbl}
\usepackage{multirow}
\usepackage{bigstrut}
\usepackage{float}
\usepackage[all]{hypcap}
\usepackage{pifont}
\usepackage{amsmath}
\usepackage{mathtools}
\usepackage{mathrsfs}
\usepackage{nicefrac}

\usepackage{algorithm}
\usepackage{algpseudocode}

\usepackage{microtype}
\usepackage{enumitem}
\usepackage{cleveref}
\usepackage{url}
\usepackage{wrapfig}
\usepackage{stackengine}
\usepackage{adjustbox}
\usepackage{rotating}
\usepackage{makecell}

\hypersetup{
  pdftitle={LongScape: Advancing Long-Horizon Embodied World Models with Context-Aware MoE},
  pdfauthor={Lei Jin, Yu Shang, Yiding Ma, Xin Zhang, Chen Gao, Wei Wu, Yong Li},
}

\begin{document}

\Cover{
  title={LongScape: Advancing Long-Horizon Embodied World Models with Context-Aware MoE},
  authors={Lei Jin\textsuperscript{*1}, Yu Shang\textsuperscript{*1}, Yiding Ma\textsuperscript{1}, Xin Zhang\textsuperscript{2}, Chen Gao\textsuperscript{1}, Wei Wu\textsuperscript{2}, Yong Li\textsuperscript{1$\dagger$}},
  affils={\textsuperscript{1}Tsinghua University \quad \textsuperscript{2}Manifold AI\\ \textsuperscript{*}Equal contribution \quad \textsuperscript{$\dagger$}Corresponding author: liyong07@tsinghua.edu.cn},
  abstract = {\textbf{Abstract.} Video-based world models hold significant potential for generating high-quality embodied manipulation data. However, current video generation methods struggle to achieve stable long-horizon generation: classical diffusion-based approaches often suffer from temporal inconsistency and visual drift over multiple rollouts, while autoregressive methods tend to compromise on visual detail. To solve this, we introduce LongScape, a hybrid framework that adaptively combines intra-chunk diffusion denoising with inter-chunk autoregressive causal generation. Our core innovation is an action-guided, variable-length chunking mechanism that partitions video based on the semantic context of robotic actions. This aligns chunk boundaries with robotic interaction events and motion dynamics, enabling the model to flexibly generate diverse dynamics. We further introduce a Context-aware Mixture-of-Experts (CMoE) framework that adaptively activates specialized experts for each chunk during generation to produce high-quality videos with coherent chunk transitions. Extensive experiments on two embodied benchmarks demonstrate that our method achieves state-of-the-art generation performance. Notably, LongScape maintains high generation quality and temporal consistency over 1,000-frame rollouts. When used as synthetic training data, LongScape videos lift the success rate of a downstream VLA policy by 9.8 percentage points over real-only training, outperforming augmentation with other long-video generators. Our code is available at: \url{https://github.com/tsinghua-fib-lab/Longscape}.},
  pageurl={https://github.com/tsinghua-fib-lab/Longscape},
  headerlogo={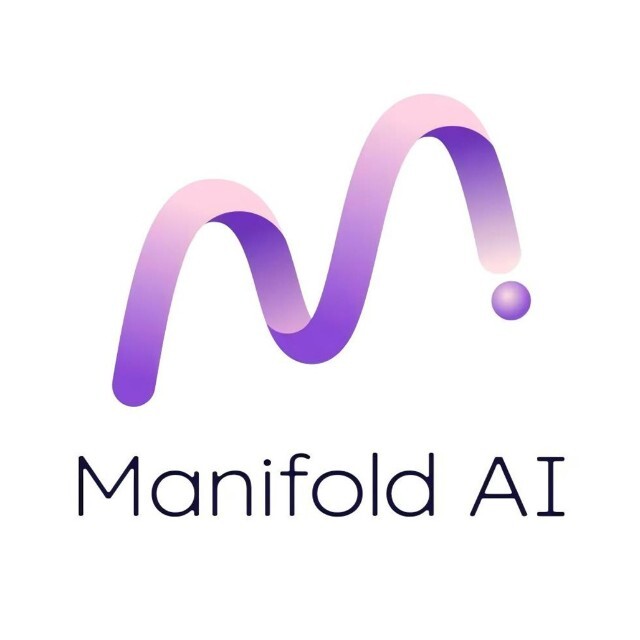}
}

\input{sec/1_introduction}

\input{sec/2_related_work}

\input{sec/3_method}

\input{sec/4_experiments}

\input{sec/5_conclusion}
\setlength{\bibsep}{0pt}
\bibliography{ref}

\input{sec/6_appendix}

\end{document}

%% file: sec/1_introduction.tex
\section{Introduction}

Video-based world models have become a prominent research direction in embodied intelligence~\citep{zhu2024irasim,liao2025genie,zhen2025tesseract,shang2025roboscape}. By learning environmental dynamics from embodied video data, these models function as powerful synthetic data engines capable of generating large-scale simulated experience to train downstream embodied policy models (such as VLA), thereby alleviating the data scarcity problem in embodied learning~\citep{jiang2025enerverse,jang2025dreamgen,agarwal2025cosmos}. 
However, current world models are predominantly limited to generating only very short video clips. For more long-horizon embodied tasks in the real world, these models struggle to provide stable generation quality, posing a major challenge to their practical deployment.

Existing video world models broadly fall into diffusion-based, autoregressive, and hybrid categories, each with inherent limitations for long-horizon generation. Diffusion models~\citep{team2025aether,liao2025genie} lack explicit causal structure and often suffer from temporal drift over extended rollouts. Autoregressive models~\citep{bruce2024genie,wu2024ivideogpt} preserve temporal causality but compromise on visual fidelity. Hybrid approaches~\citep{deng2024autoregressive,teng2025magi,zhuang2025video} combine diffusion denoising with autoregressive generation, yet existing methods rely on fixed-length chunks that arbitrarily segment semantically continuous actions or conflate distinct motion patterns (e.g., locomotion and manipulation) into a single chunk. This semantic ambiguity misguides next-chunk prediction and degrades long-range temporal coherence.

To address these challenges, we introduce \textbf{LongScape}, an embodied world model designed for long-horizon video generation. Our framework adopts a hybrid approach combining intra-chunk diffusion denoising with inter-chunk autoregressive causal generation. The key innovation is a context-adaptive, variable-length chunking mechanism guided by robot action priors: frames with gripper state changes or substantial effector motion are partitioned into shorter chunks for fine-grained manipulation, while frames with minimal motion form longer chunks for efficient generation. We pre-process embodied datasets with this action-based rule to create chunks of four distinct granularities, and train multiple Diffusion Transformer (DiT) experts, each specializing in a specific temporal dynamic (e.g., fine manipulation vs.\ locomotion). Unlike conventional MoE architectures~\cite{riquelme2021scaling,li2025uni} that emphasize load balancing, our experts are explicitly specialized by task context. A context-based dynamic router, conditioned on text instructions and current visual features, selects the appropriate expert at each rollout step, ensuring semantic coherence and causal consistency throughout the generated sequence.

We conduct extensive experiments on the LIBERO~\citep{liu2023libero} and AGIBOT-World~\citep{bu2025agibot} benchmarks. Our model achieves state-of-the-art performance in video generation quality compared to existing diffusion, autoregressive, and hybrid baselines, maintaining visual coherence and stability over 50 rollouts and 1000 frames. Our primary contributions are:

\begin{itemize}[leftmargin=*]
    \item We propose LongScape, a novel embodied world model with adaptive chunk rollout that enables stable long-horizon video generation.
    \item We introduce an action-prior-guided chunk partitioning scheme that promotes semantic coherence of video chunks for effective causal autoregressive generation.
    \item Extensive experiments demonstrate state-of-the-art long-horizon generation quality with sustained stability in visual quality and motion correctness.
\end{itemize}

%% file: sec/2_related_work.tex
\section{Related Works}

\subsection{Video Generation Models}
Video generation has been approached through diffusion models, autoregressive models, and hybrid frameworks.
Diffusion models~\citep{ho2022video,yang2024cogvideox,kong2024hunyuanvideo,wan2025wan} iteratively denoise latent representations and achieve high visual fidelity but lack explicit temporal causality, leading to drift over long sequences. 
Flow-based video world models such as Cosmos-Predict2.5~\citep{nvidia2025worldsimulationvideofoundation} use a DiT architecture trained with rectified flow. Autoregressive models~\citep{bruce2024genie,wu2024ivideogpt} formulate generation as next-token prediction, which preserves causality but often yields lower visual quality. 
Hybrid models~\citep{deng2024autoregressive,teng2025magi} combine diffusion denoising within an autoregressive chain to balance both aspects. 
However, existing hybrid methods employ fixed-length chunks that ignore the varying temporal dynamics of embodied tasks, where manipulation and locomotion demand different temporal granularities.

\subsection{Long-Horizon Video Generation}
Long-video generation approaches can be grouped into divide-and-conquer strategies (e.g., hierarchical keyframe planning~\citep{Wang_2024_CVPR,zhao2025moviedreamerhierarchicalgenerationcoherent,yin-etal-2023-nuwa}) and temporal autoregressive methods that generate chunk-by-chunk~\citep{gao2024vidgptintroducinggptstyleautoregressive,teng2025magi,zhuang2025video}. 
Both families share a common limitation: fixed temporal granularity. This rigidity prevents adaptation to the heterogeneous motion patterns inherent in embodied tasks.
Our LongScape framework introduces adaptive variable-length chunking driven by action priors, applying fine-grained diffusion to short chunks during critical interactions and efficient generation to longer chunks during stable periods.

\subsection{Embodied World Models}
Embodied world models learn environmental dynamics to support synthetic data generation and policy learning~\citep{ding2024understanding}. 
They broadly follow two paradigms: video generation-based methods that adapt foundation models for text- or action-conditioned synthesis~\citep{wu2024ivideogpt,liao2025genie,huang2025enerverse,jiang2025enerverse,zhen2025tesseract,shang2025roboscape}, and latent prediction methods that forecast compact state representations for planning~\citep{assran2025v,baldassarre2025back}.
Our work belongs to the text-conditioned video generation paradigm, with a focus on enabling stable long-horizon synthesis for embodied scenarios.
We note that LongScape is conceptually adjacent to, but distinct from, action-conditioned embodied world models such as EnerVerse-AC~\citep{jiang2025enerverse} and HMA~\citep{wang2025hma}. These models ingest per-step robot actions and serve as interactive predictors for closed-loop control and policy evaluation, whereas LongScape is a text-conditioned generative model aimed at offline synthesis of long-horizon training data, and does not consume action inputs at inference time. Because the conditioning signals, use cases, and evaluation protocols differ substantially, we treat them as complementary rather than direct baselines; extending LongScape with explicit action conditioning is a promising future direction.

%% file: sec/3_method.tex
\section{Methodology}

\begin{figure*}[t]
\begin{center}
\includegraphics[width=0.99\textwidth]{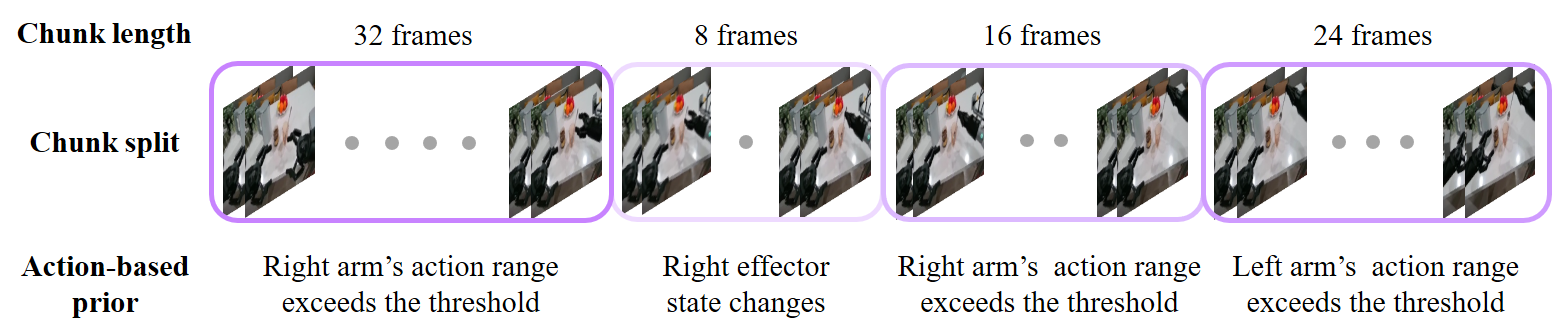}
\end{center}
\caption{Action-guided chunk partitioning mechanism. This mechanism leverages embodied actions to partition the video, ensuring each chunk contains a distinct semantic unit. It divides the stream into short segments for active manipulation (e.g., significant movement or end-effector state changes) and longer segments for subtle locomotion dynamics.}
\label{fig:partition}
\end{figure*}

\subsection{Autoregressive video generation with chunk-wise diffusion}

Embodied world models broadly fall into two paradigms: action-driven models for online interactive control, and text-driven models for offline data synthesis. LongScape belongs to the latter, targeting stable long-horizon text-conditioned generation as a foundation for embodied data augmentation (validated in Section~\ref{sec:policy_learning}).

Inspired by the next-token-prediction autoregressive paradigm of LLMs, we approach video generation by treating a video as a sequence of temporal chunks, with each chunk (single or multiple frames) functioning as a ``token''. This enables a token-by-token rollout to generate videos, a causal paradigm that inherently preserves causality and helps enforce physical consistency. The full video can be represented as $\mathbf{V} = (\mathbf{S}_1, \mathbf{S}_2, \dots, \mathbf{S}_N)$, where $\mathbf{S}_t$ is the video chunk with a timestamp $t$. The objective is to predict the next chunk based on the preceding history: 
\begin{equation}
    p(\mathbf{V}) = p(\mathbf{S}_1) \prod_{t=1}^{N-1} p(\mathbf{S}_{t+1} | \mathbf{S}_1, \dots, \mathbf{S}_t),
\end{equation}
where $\mathbf{S}_t$ is the video chunk at time step $t$. 
The central task then becomes modeling the single-step generation process $p(\mathbf{S}_{t+1} | \mathbf{S}_1, \dots, \mathbf{S}_t)$. Two primary paradigms exist to address this.

The first is to handle visual information similarly to language, using a tokenizer to discretize the image into patches and then extracting discrete representations. These are then arranged in spatiotemporal order to form the chunk's overall token: $\mathbf{X}_t$, where $\mathbf{X}_t = \text{Tokenizer}(\mathbf{S}_t)$. A transformer-based model is then trained on a next-token prediction task to predict the subsequent chunk token $\mathbf{X}_{t+1}$. While this approach is generally efficient, the discretization of continuous visual information can lead to a reduction in generated image quality. 

The second paradigm is diffusion-based continuous generation, where the next video chunk, $\mathbf{S}_{t+1}$, is produced through a conditional denoising process guided by the previous chunk, $\mathbf{S}_t$. Operating within a continuous latent space (often from a VAE), the model starts with a noisy latent representation of the target chunk, $\mathbf{z}_{t+1, K}$. This representation is then iteratively refined over $K$ steps.
At each denoising step $k$, a model, typically a U-Net or a Diffusion Transformer (DiT), predicts and removes noise from the current noisy sample, $\mathbf{z}_{t+1, k}$. This operation is conditioned on the preceding chunk $\mathbf{S}_t$, following the process: $\mathbf{z}_{t+1, k-1} = \text{Denoise}(\mathbf{z}_{t+1, k}, \mathbf{S}_t)$. This refinement ultimately yields a clean latent representation, $\mathbf{z}_{t+1, 0}$, which is then decoded into the final video chunk, $\mathbf{S}_{t+1}$. This approach preserves finer local visual details, mitigates error accumulation and drift during long-range autoregressive rollout, and benefits from the powerful generative capabilities of modern full-sequence diffusion models.
Given these advantages, we adopt a hybrid framework that combines local chunk-based diffusion denoising with global autoregressive generation. In the following section, we elaborate on our design of adaptive chunk partitioning to enable more efficient long-horizon video generation within this framework.

\subsection{Action-guided video chunk partition}
Our design for the  Context-aware Mixture-of-Experts (CMoE) in embodied video generation stems from the observation that different phases of a robotic task involve distinct dynamic patterns. For instance, long-range navigation toward an object requires the model to capture locomotion patterns, while precise manipulation demands an understanding of fine-grained dynamics. 
Accordingly, we propose to partition video sequences using action-state changes as a structural prior.

We focus on embodied manipulation scenarios, where robotic actions are naturally characterized by the gripper state and end-effector pose, where the gripper state indicates the interaction events, while the end-effector pose captures motion magnitude and direction. 
Concretely, we posit that robotic action labels provide a reliable signal for video chunk partitioning, based on two key factors: (1) changes in the gripper state reveal whether the robot is interacting with an object, and (2) motion magnitude can be quantified with end-effector pose changes. The video is first divided into non-overlapping base chunks of 8 frames. Each base chunk in which the gripper state changes is then isolated as an independent chunk, so that every interaction event is preserved at the finest temporal granularity. The remaining segments are partitioned by grouping 1 to 4 consecutive base chunks, resulting in variable-length chunks of 8, 16, 24, or 32 frames.

For each action dimension $d$, we define the motion threshold as $\theta_d = \alpha R_d$, where $R_d = \max_f A_d(f) - \min_f A_d(f)$ is the motion range over the full video and $\alpha=0.5$ is the default threshold ratio. The partitioning algorithm processes the video sequentially from the initial frame. It first partitions the video immediately before and after every base chunk in which the gripper state changes, isolating each of them as an independent chunk. Each remaining segment is then processed in order: starting with its first base chunk, the algorithm iteratively evaluates a candidate chunk consisting of the next \(n\) consecutive base chunks (\(1 \leq n \leq 4\)). The partitioning steps proceed as follows:

\begin{enumerate}[leftmargin=*]
    \item \textbf{Substantial Motion Detection:} If the range of motion in any dimension within the candidate chunk exceeds \(\theta_d\) (\(\theta_d = \alpha R_d\)), all \(n\) base chunks are grouped into a single chunk.
    \item \textbf{Maximum Length Check:} If \(n = 4\), the candidate chunk is immediately assigned as a chunk to avoid exceeding the maximum allowed length.
    \item \textbf{Segment End Check:} If the candidate chunk reaches the end of the current segment, it is assigned as the final chunk of the segment.
    \item \textbf{Continue Expansion:} If none of the above conditions are met, \(n\) is incremented by 1, and the process repeats—unless \(n\) already equals 4.
\end{enumerate}

This approach ensures that video segments with significant motion or critical gripper state changes are assigned shorter chunks, while segments with less dynamic activity are assigned longer chunks. The overall process is formalized in Algorithm~\ref{alg:chunk}.

\begin{figure}[t]
\begin{center}
\includegraphics[width=0.9\textwidth]{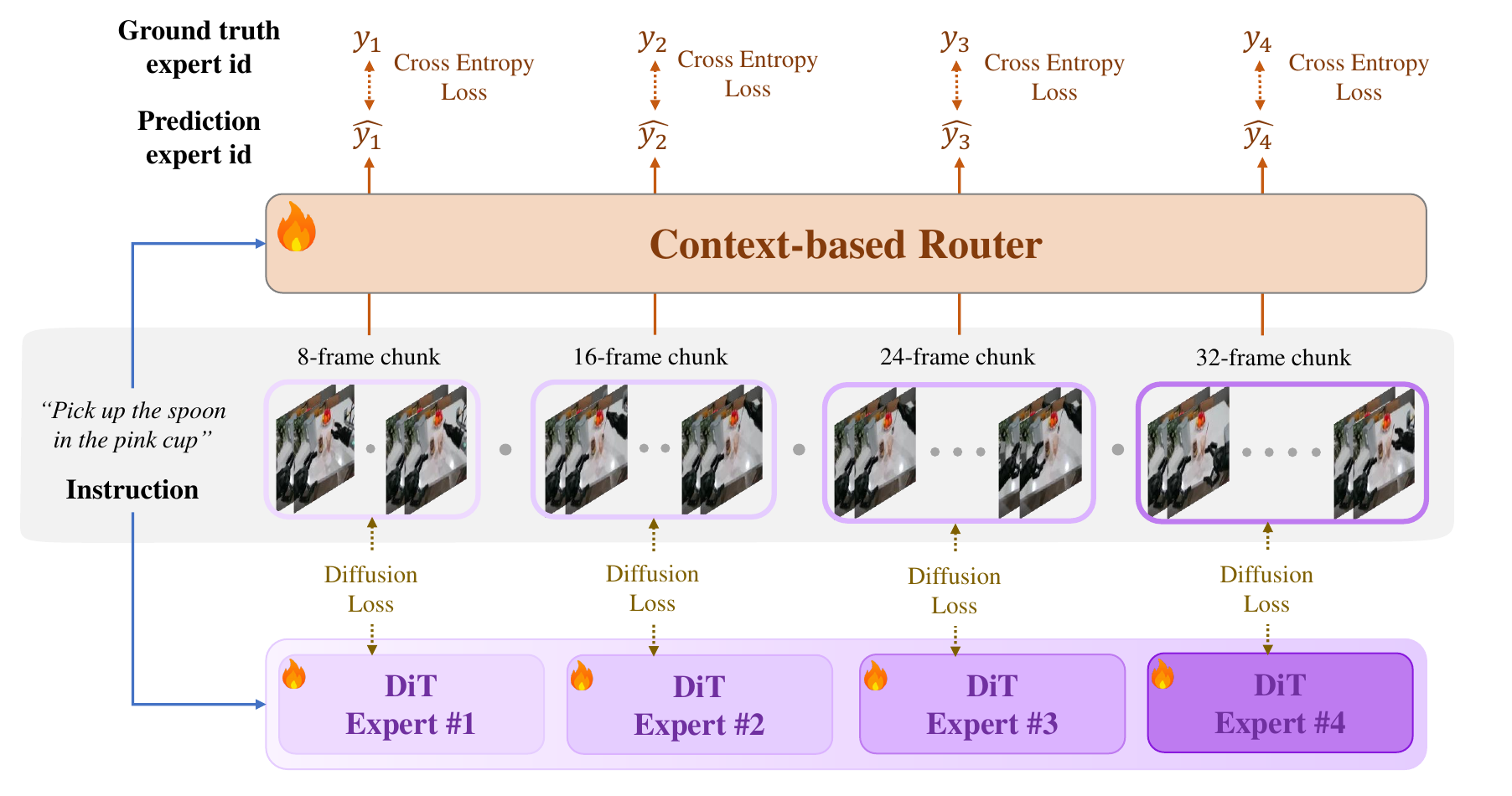}
\end{center}
\caption{Training paradigm of LongScape. Different types of video chunks are used to train specialized DiT experts, while a context-based dynamic router learns to allocate the next chunk to the appropriate expert based on textual instruction and visual context.}
\label{fig:train}
\end{figure}

\subsection{Context-aware MoE for adaptive rollout}
Following the partitioning of video into chunks of varying lengths, we introduce a \textbf{Context-aware Mixture-of-Experts} \textbf{(CMoE)} framework to reconcile specialized DiT networks for denoising each chunk type. This approach mitigates conflicts and forgetting issues that arise from training on diverse motion patterns, e.g., large-scale locomotion versus fine-grained object manipulation.
Our model employs a set of $K$ DiT experts, denoted as $\mathcal{E} = \{ \mathcal{E}_1, \mathcal{E}_2, \dots, \mathcal{E}_K \}$. Each expert is initialized from the parameters of a pre-trained CogVideoX~\citep{yang2024cogvideox}. 
In our implementation, we set $K=4$, corresponding to four distinct chunk types. Each expert network is composed of a stack of transformer blocks, featuring 3D full attention layers and feed-forward networks connected by Adaptive Layer Normalization (AdaLN) layers. We train each expert $\mathcal{E}_i$ to perform a denoising task on its corresponding chunk type, optimizing the objective function:
\begin{equation}
    \mathcal{L}_i = \mathbb{E}_{\mathbf{z}_0, \epsilon \sim \mathcal{N}(0, \mathbf{I}), \tau \sim [1, T]} \left[ w_\tau \left\| \mathbf{v}_\tau - \mathcal{E}_i(\mathbf{z}_\tau, \tau, \mathbf{c}, \mathbf{S}_t) \right\|^2 \right],
\end{equation}
where $\mathbf{z}_0$ is the clean latent representation of the target chunk $\mathbf{S}_{t+1}$, $\epsilon$ is the added noise, $\mathbf{z}_\tau = \sqrt{\bar{\alpha}_\tau}\, \mathbf{z}_0 + \sqrt{1-\bar{\alpha}_\tau}\, \epsilon$ is the noisy latent at diffusion time step $\tau$, $\bar{\alpha}_\tau$ is the cumulative noise schedule, $T$ is the total number of diffusion steps, $\mathbf{c}$ is the global text condition, and $\mathbf{S}_t$ is the preceding video chunk. The training target is the velocity $\mathbf{v}_\tau = \sqrt{\bar{\alpha}_\tau}\, \epsilon - \sqrt{1-\bar{\alpha}_\tau}\, \mathbf{z}_0$, and $w_\tau = 1/(1-\bar{\alpha}_\tau)$ is the per-timestep loss weight used in our implementation. Since each expert is initialized from CogVideoX, we retain this velocity parameterization during fine-tuning so that the optimization objective remains consistent with the pretrained model.

To enable adaptive inference-time expert selection, we introduce a \textbf{dynamic contextual router} $\mathcal{R}$, which is implemented by a single cross-attention transformer network, taking the global text instruction $\mathbf{c}$ and visual features from the current video chunk $\mathbf{S}_t$ as input, predicting which expert to activate for the next chunk $\mathbf{S}_{t+1}$.
The router is trained using a cross-entropy loss:
\begin{equation}
    \mathcal{L}_{\text{router}} = \mathbb{E}_{(\mathbf{c}, \mathbf{S}_t, i)} \left[ -\log p(i | \mathbf{c}, \mathbf{S}_t) \right],
\end{equation}
where $i$ is the ground-truth expert index.
The training process of experts and the router is presented in Figure~\ref{fig:train} and this process is highly efficient, requiring only a small subset of partitioned video chunks and their corresponding chunk length labels. 

\begin{figure}[t]
\begin{center}
\includegraphics[width=0.75\textwidth]{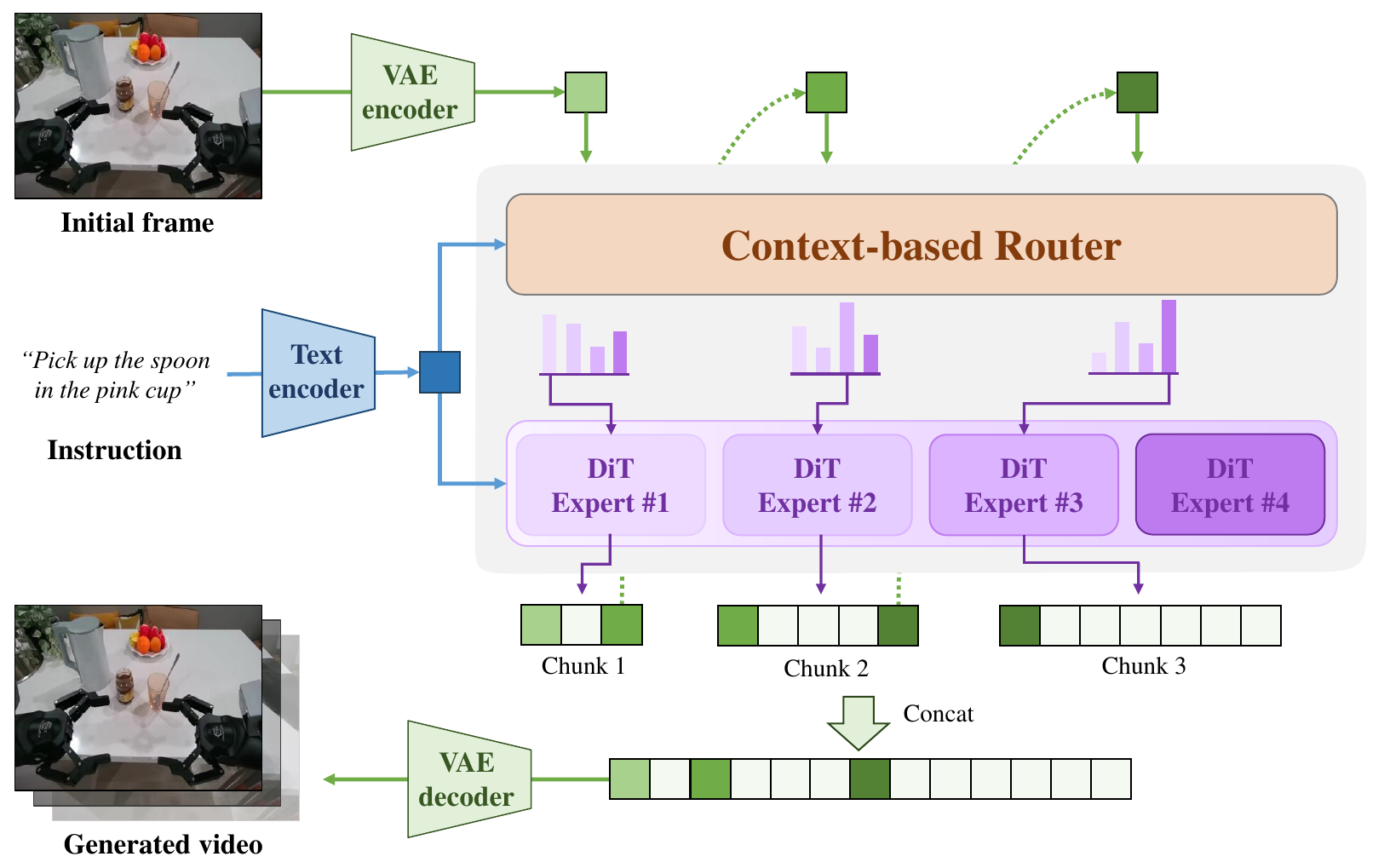}

\end{center}
\caption{Inference pipeline of LongScape. At each autoregressive step, the dynamic router leverages the global text instruction and the visual features of the current chunk to select the appropriate DiT expert. The selected expert then generates the subsequent video chunk via diffusion denoising, with this iterative process continuing to produce a long video sequence.}
\label{fig:infer}
\end{figure}

The inference pipeline is illustrated in Figure~\ref{fig:infer}. At each rollout step, the global text prompt and visual features from the last frame of the current chunk are fed into the router to predict the appropriate expert:
\begin{equation}
    i^* = \underset{i \in \{1, \dots, K\}}{\arg \max} \left( \mathcal{R}(\mathbf{c}, \mathbf{S}_t)_i \right).
\end{equation}
The activated expert $\mathcal{E}_{i^*}$ then generates the next chunk of the corresponding length $\mathbf{S}_{t+1}$:
\begin{equation}
    \mathbf{S}_{t+1} = \mathcal{E}_{i^*}(\mathbf{S}_t, \mathbf{c}).
\end{equation}
We note that action labels (gripper state and end-effector pose) are consumed only \emph{offline}, for partitioning training videos into chunks and supervising expert specialization; at inference time, LongScape requires only the text instruction and the visual features of the current chunk, and never queries action annotations.

Although the current design focuses on robotic manipulation scenarios, the proposed CMoE framework is generally applicable to long-horizon video generation with structured dynamics, and can be extended to other action-annotated domains such as mobile navigation and human motion. We leave large-scale validation on general long-form video benchmarks to future work.

\begin{table*}[t]
\centering
\caption{Quantitative comparison of different video world models in terms of video visual quality on LIBERO and AGIBOT-World datasets.}
\label{tab:result}
\resizebox{\textwidth}{!}{
\begin{tabular}{ccccccccccccc}
\toprule
\multirow{2}{*}{\textbf{Method}} & \multicolumn{6}{c}{\textbf{LIBERO}} & \multicolumn{6}{c}{\textbf{AGIBOT-World}} \\
\cmidrule(lr){2-7} \cmidrule(lr){8-13}
 & PSNR$\uparrow$ & LPIPS$\downarrow$ & SSIM$\uparrow$ & FVD$\downarrow$ & \makecell{Subject\\ Consistency$\uparrow$} & \makecell{Background \\Consistency$\uparrow$} & PSNR$\uparrow$ & LPIPS$\downarrow$ & SSIM$\uparrow$ & FVD$\downarrow$ & \makecell{Subject\\ Consistency$\uparrow$} & \makecell{Background \\Consistency$\uparrow$} \\
\midrule
\makecell{CogVideoX} & 19.315 & 0.1402 & 0.7729 & 184.69 & 0.8974 & 0.9582 & 15.800 & 0.4028 & 0.6733 & 267.39 & 0.8224 & 0.8874\\
\makecell{Genie} & 19.300 & 0.2192 & 0.7267 &383.82 & 0.9098 & 0.9592 & 16.166 & 0.3620 & 0.6058 & 480.45 & 0.8743 & 0.9180\\
\makecell{Cosmos-Predict2.5} & 10.047 & 0.5005 & 0.1587 &413.53 & 0.7068 & 0.8832 & 12.523 & 0.4212 & 0.5331 & 268.71 & 0.8597 & 0.9093\\
\makecell{NOVA} & 13.007 & 0.4488 & 0.4262 & 346.91 & 0.9114 & 0.9580 & 16.043 & 0.4201 & 0.6998 & 273.60 & 0.8148 & 0.9018\\
\makecell{MAGI-1} & 15.330 & 0.3520 & 0.4721 &541.39 & 0.8847 & 0.9353 & 14.448 & 0.4849 & 0.5961 & 490.53 & 0.9092 & 0.9265\\
\makecell{CausVid} & 19.915 & 0.1383 & 0.7860 &276.25 & 0.9160 & 0.9566 & 15.873 & 0.4934 & 0.6920 & 358.33 & 0.6965 & 0.8378\\
\makecell{Genie-Envisioner} & 18.204 & 0.2832 & 0.6848 &417.55 &0.9175 &0.9569 & 14.468 & 0.5193 & 0.6323 & 495.93 &0.9054 & 0.9192\\
\makecell{\textbf{Ours}} & \textbf{19.977} & \textbf{0.1231} & \textbf{0.7883} & \textbf{153.72}& \textbf{0.9297} & \textbf{0.9622} &\textbf{16.493} & \textbf{0.3613} & \textbf{0.7015} & \textbf{256.16} & \textbf{0.9207} & \textbf{0.9341}\\
\bottomrule
\end{tabular}
}
\end{table*}

\begin{figure*}[t]
\begin{center}
\includegraphics[width=\textwidth]{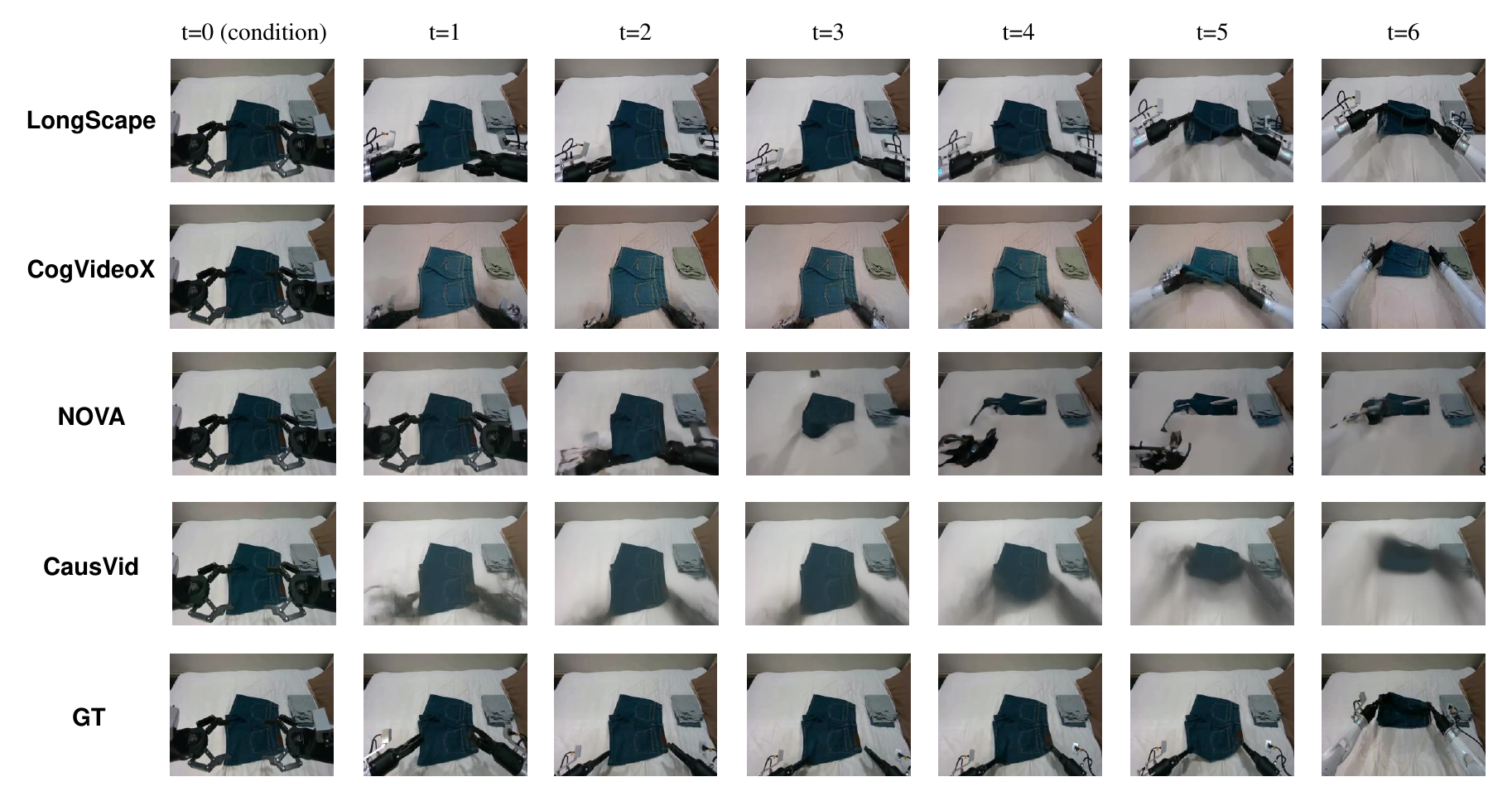}

\end{center}
\caption{Qualitative comparison on AGIBOT-World dataset with text instruction ``grasp the lower pant leg and waistband of denim shorts and fold them over the upper pant leg and waistband''. Here we select several models with good results and sample six frames from the generated long video sequences for better presentation.}
\label{fig:main_agi}
\end{figure*}

%% file: sec/4_experiments.tex
\section{Experiments}

\subsection{Experimental Setup} \label{sec:4.1}

\textbf{Datasets.} We conduct experiments on LIBERO~\citep{liu2023libero} and AGIBOT-World~\citep{bu2025agibot}, both focusing on long-horizon composite embodied manipulation tasks (e.g., folding clothes). Detailed descriptions are provided in Appendix~\ref{appendix:dataset}.

\textbf{Baselines.} We compare LongScape against three categories of state-of-the-art models:
\begin{itemize}[leftmargin=*]
    \item \textbf{Foundational Video Generation Models:} CogVideoX~\citep{yang2024cogvideox}, Genie~\citep{bruce2024genie}, Cosmos-Predict2.5~\citep{nvidia2025worldsimulationvideofoundation}, and NOVA~\citep{deng2024autoregressive}.
    \item \textbf{Long-Horizon Video Generation Models:} MAGI-1~\citep{teng2025magi} and CausVid~\citep{zhuang2025video}.
    \item \textbf{Embodied World Models:} Genie-Envisioner~\citep{liao2025genie}.
\end{itemize}
This selection covers both general video generation architectures and specialized long-horizon methods. Further model details are in Appendix~\ref{appendix:baseline}.

\textbf{Metrics.} We evaluate using PSNR, SSIM, LPIPS, FVD, subject consistency, and background consistency (see Appendix~\ref{appendix:metrics}).

\textbf{Implementation.} Each DiT expert has 5.57B parameters; the router has 108.57M (AGIBOT-World) and 71.35M (LIBERO) parameters. All models are trained on 30,000 clips sampled at 10\,Hz. Full training configurations and resource requirements are detailed in Appendix~\ref{appendix:implementation}.

\subsection{Main results}

\begin{figure*}[t]
\begin{center}
\includegraphics[width=\textwidth]{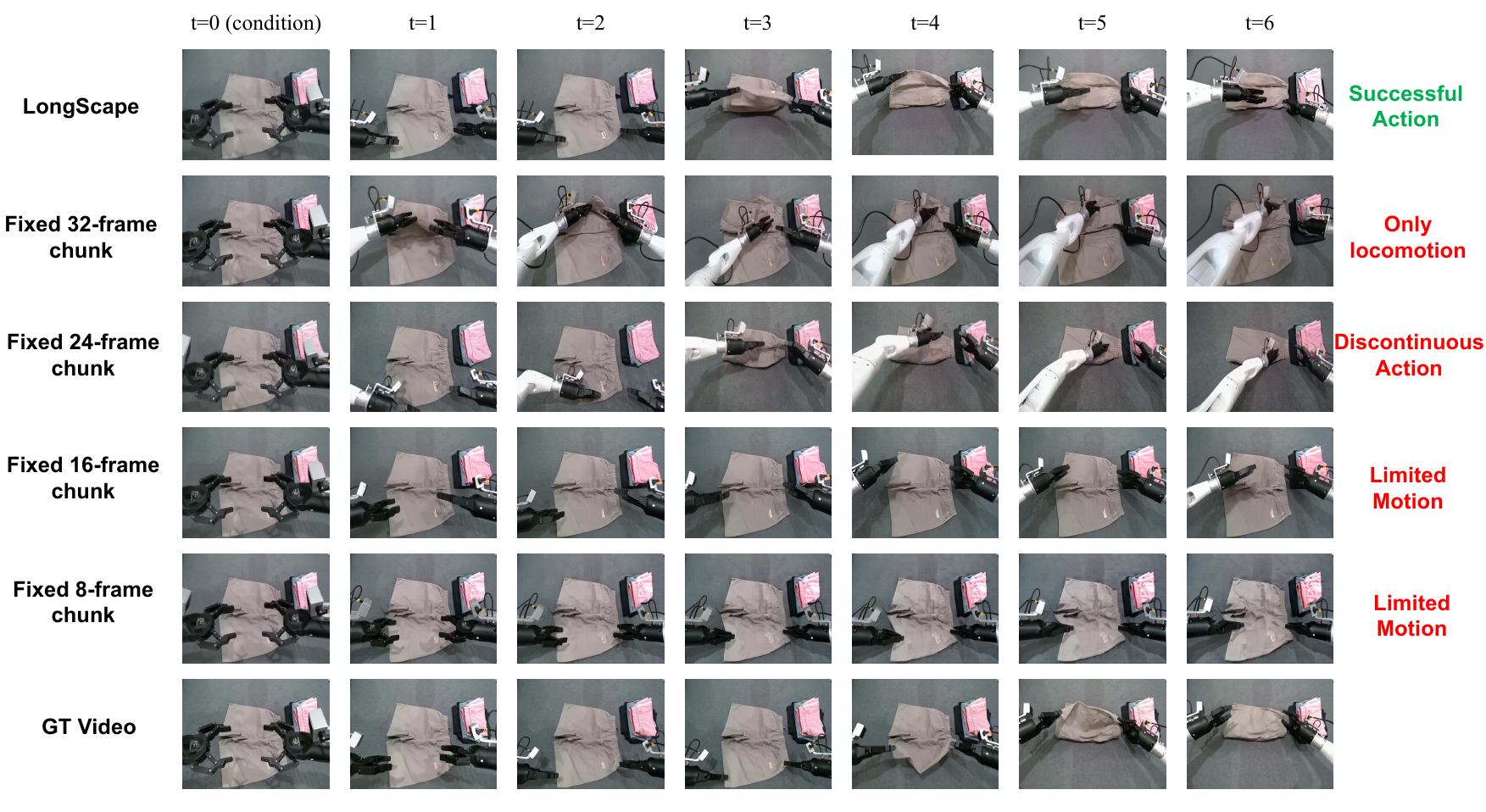}

\end{center}
\caption{Visualization of the ablation study on AGIBOT-World dataset. We compare LongScape with variants using only fixed-length chunks. These fixed-length approaches often exhibit discontinuous or physically implausible actions. In contrast, LongScape, by incorporating an adaptive expert mechanism, ensures action coherence and stable generation.}
\label{fig:ab_agi}
\end{figure*}

\begin{table*}[t]
\centering
\caption{Ablation study of using various fixed-length chunk generation and the MoE designs.}
\label{tab:ablation}
\resizebox{\textwidth}{!}{
\begin{tabular}{ccccccccccccc}
\toprule
\multirow{2}{*}{\textbf{Method}} & \multicolumn{6}{c}{\textbf{LIBERO}} & \multicolumn{6}{c}{\textbf{AGIBOT-World}} \\
\cmidrule(lr){2-7} \cmidrule(lr){8-13}
 & PSNR$\uparrow$ & LPIPS$\downarrow$ & SSIM$\uparrow$ & FVD$\downarrow$ & \makecell{Subject\\ Consistency$\uparrow$} & \makecell{Background \\Consistency$\uparrow$} & PSNR$\uparrow$ & LPIPS$\downarrow$ & SSIM$\uparrow$ & FVD$\downarrow$ & \makecell{Subject\\ Consistency$\uparrow$} & \makecell{Background \\Consistency$\uparrow$} \\
\midrule
\textbf{LongScape}  & \textbf{19.977} & \textbf{0.1231} & \textbf{0.7883} & \textbf{153.72} & \textbf{0.9297} & \textbf{0.9622}& \textbf{16.493} & \textbf{0.3613} & \textbf{0.7015} & \textbf{256.16} & \textbf{0.9207} & \textbf{0.9341}\\
32-frame chunk & 19.755 & 0.1285 & 0.7845 & 180.31 & 0.9091 & 0.9610 & 15.641 & 0.4286 & 0.6702 & 282.38 & 0.8535 & 0.9111 \\
24-frame chunk& 19.620 & 0.1314 & 0.7798 & 191.02 & 0.9104 & 0.9620 & 15.903 & 0.4034 & 0.6814 & 303.75 & 0.8952& 0.9328 \\
16-frame chunk & 19.640 & 0.1388 & 0.7783 & 176.55 & 0.9129 & 0.9614 & 15.894 & 0.3852 & 0.6917 & 373.50 & 0.9090 & 0.9328 \\
8-frame chunk& 18.893 & 0.1629 & 0.7518 & 229.71 & 0.8999 & 0.9599 & 15.578 & 0.4052 & 0.6669 & 387.49 & 0.8789& 0.9178\\
Unified Expert& 12.895 & 0.4277 & 0.4387 & 365.35 & 0.9028 & 0.9583 & 14.894 & 0.4595 & 0.6532 & 389.24 & 0.8713 & 0.9103 \\
\bottomrule
\end{tabular}
}
\end{table*}

\textbf{Quantitative comparison.}
Quantitative evaluations in Table~\ref{tab:result} demonstrate that our model achieves superior performance across all six metrics on both the LIBERO and AGIBOT-World datasets. These benchmarks focus on \textbf{long-horizon composite embodied manipulation tasks} (each task lasting approximately 20 seconds), which generally require multiple rollout steps to generate the full video sequence.
We further evaluate LongScape over 1,000-frame rollouts on AGIBOT-World, demonstrating sustained generation quality and temporal consistency across the test rollouts (Appendix~\ref{sec:longhorizon}).
Compared with CogVideoX, LongScape achieves average relative improvements of 8.6\% on LIBERO and 5.8\% on AGIBOT-World across PSNR, LPIPS, SSIM, and FVD. We compute the arithmetic mean of the relative gains over CogVideoX, using $(m_{\mathrm{ours}}-m_{\mathrm{base}})/m_{\mathrm{base}}$ for PSNR and SSIM and $(m_{\mathrm{base}}-m_{\mathrm{ours}})/m_{\mathrm{base}}$ for LPIPS and FVD.
Our baseline suite is organized by generation paradigm: CogVideoX represents pure full-sequence diffusion, Genie represents pure autoregressive prediction, and NOVA, MAGI-1, and CausVid represent hybrid diffusion--autoregressive methods of the same class as LongScape. Each baseline is evaluated under its native inference protocol: diffusion models generate the entire long frame sequence in a single forward pass, while autoregressive and hybrid models produce long sequences through iterative rollouts. Converting CogVideoX into a chunk-wise autoregressive generator would no longer evaluate the pure diffusion paradigm this comparison is designed to probe, and LongScape's margin remains clear against the autoregressive and hybrid baselines that share the same rollout-style evaluation.
However, these methods have significant drawbacks.
Diffusion models often generate physically implausible content due to their parallel frame generation mechanism, while the autoregressive models suffer from error accumulation, where early inaccuracies propagate and amplify over time. 
These limitations hinder both baseline types from producing coherent long-range videos.
Our model's architecture avoids these issues by flexibly generating chunks with full semantic integrity. This approach ensures high-quality generation within each chunk and maintains strong causal relationships between them, making our model exceptionally well-suited for long-horizon generation tasks.

\textbf{Qualitative results.}
To provide a more intuitive comparison of our method's effectiveness against the baselines, we offer a set of visualizations of the generated results.
Figure~\ref{fig:main_agi} presents a challenging cloth-folding task from the AGIBOT-World dataset. This task requires a high level of spatiotemporal coherence and a strong understanding of physical laws. Here, CogVideoX produces blurry artifacts. NOVA suffers from objects disappearing from the scene. By comparison, our model consistently maintains a plausible deformation of the cloth and generates a continuous, coherent motion sequence even with multiple rollouts.
Figure~\ref{fig:main_libero} shows the generation of a complex manipulation task from the LIBERO dataset. This task involves multiple steps, such as picking and placing objects and closing a drawer. 

As shown in the visualization, CogVideoX suffers from ghosting artifacts in later frames. CausVid again blurs subject. NOVA misinterprets the instruction, picking up the wrong object, while Genie fails in the very first pickup action. 
In contrast, our model maintains coherent actions across multiple rollouts, producing a logical and successfully executed task sequence.

\begin{table}[t]
\centering
\caption{Per-frame generation time (in seconds) on LIBERO dataset.}
\label{tab:throughput}
\resizebox{0.7\textwidth}{!}{
\begin{tabular}{cccccc}
\toprule
\textbf{Method} & \makecell{32-frame\\chunk}& \makecell{24-frame\\chunk} & \makecell{16-frame\\chunk} & \makecell{8-frame\\chunk} & LongScape \\
\midrule
\textbf{Time/Frame (s)} & 0.5598 & 0.6189 & 0.7323 & 1.0912 & 0.7974\\
\bottomrule
\end{tabular}}
\end{table}

\begin{table}[t]
\centering
\caption{Quantitative comparison of different models on out-of-domain tasks on AGIBOT-World dataset. }
\label{tab:ood}
\resizebox{0.75\textwidth}{!}{
\begin{tabular}{ccccccc}
\toprule
\textbf{Method} & PSNR$\uparrow$ & LPIPS$\downarrow$ & SSIM$\uparrow$ & FVD$\downarrow$ & \makecell{Subject\\ Consistency$\uparrow$} & \makecell{Background \\Consistency$\uparrow$} \\
\midrule
CogVideoX & 15.528 & 0.3877 & 0.6465 & 301.80 & 0.8649 & 0.8916 \\
Genie & 11.863 & 0.6234 & 0.2716 & 385.78 & 0.8186 & 0.9028 \\
NOVA & 13.167 & 0.5406 & 0.4601 & 435.50 & 0.7816 & 0.8895 \\
CausVid & 9.833 & 0.6754 & 0.2376 & 470.38 & 0.7585 & 0.8449 \\
\textbf{LongScape}  & \textbf{15.936} & \textbf{0.3717} & \textbf{0.6585} & \textbf{297.16} & \textbf{0.9172} & \textbf{0.9167} \\
\bottomrule
\end{tabular}}
\end{table}

\subsection{Ablation Study}
\textbf{Effectiveness of the context-based MoE.}
To validate our MoE design, we compare LongScape against four variants each using a single expert trained on fixed-length chunks (8, 16, 24, or 32 frames). As shown in Table~\ref{tab:ablation}, all fixed-length variants underperform our CMoE on both datasets. Longer chunks (e.g., 32 frames) perform slightly better than shorter ones (e.g., 8 frames), which we attribute to very short chunks failing to capture semantically complete actions.

To evaluate the benefit of length-specific specialization under the same training data and active DiT parameter count, we compare against a \textbf{Unified Expert} baseline: a single DiT trained on the union of the same action-guided chunks used to train the four specialized experts. Each specialized expert receives only chunks of its designated length, whereas the Unified Expert learns from all four lengths jointly. Both approaches activate a single 5.57B-parameter DiT per rollout step. The Unified Expert achieves substantially worse performance (e.g., 12.895 vs.\ 19.977 PSNR on LIBERO, and 14.894 vs.\ 16.493 PSNR on AGIBOT-World) and even underperforms the best fixed-length expert on both benchmarks. These results demonstrate that, with the same training data and active DiT parameter count, length-specific expert learning outperforms mixed-length learning in a single DiT on both benchmarks.

Qualitative results (Figure~\ref{fig:ab_agi} and Figure~\ref{fig:ab_libero}) further illustrate that fixed-length designs suffer from mode collapse: long-chunk models produce only coarse locomotion, while short-chunk models exhibit poor motion dynamics. Our CMoE flexibly handles both patterns, producing consistently coherent generation.

\textbf{Effectiveness of the dynamic router.}
We evaluate the router as a four-class classification task on 12,700 LIBERO test samples, achieving 91.4\% accuracy. Despite occasional routing errors, LongScape maintains high temporal consistency over 1,000-frame rollouts, as shown in Appendix~\ref{sec:longhorizon}. With the same expert bank, the learned router outperforms random, round-robin, and visual-change-based length selection on LIBERO (Appendix~\ref{sec:length_policy}).

We further verify the robustness of our action-guided chunking by varying the threshold ratio $\alpha$ in $\theta_d = \alpha R_d$. For $\alpha \in \{0.3, 0.5, 0.7\}$, LongScape consistently outperforms the best fixed-length expert (details in Appendix~\ref{sec:threshold}). We use $\alpha = 0.5$ by default.

\textbf{Throughput analysis of the MoE architecture.}
Table~\ref{tab:throughput} reports per-frame generation time on LIBERO. Fixed-chunk models become slower as chunk length decreases; LongScape achieves 0.7974s per frame, an intermediate value balancing fine-grained detail and parallel efficiency. Training uses data-parallel fine-tuning (total compute equals one full fine-tune). The four experts together require approximately 42\,GB of storage in half precision. During inference, CPU offloading keeps inactive experts in CPU memory and loads only the selected expert onto the GPU, enabling inference on a single NVIDIA RTX 4090D GPU.

\textbf{Out-of-Domain Generalization on Unseen Tasks.}
We evaluate generalization by training exclusively on ``cloth folding'' and testing on ``pick-and-place'' in an unseen supermarket environment. As shown in Table~\ref{tab:ood}, LongScape outperforms all baselines across all metrics, confirming its ability to learn task-agnostic dynamics.

\subsection{Generated Data for Downstream Policy Learning}\label{sec:policy_learning}

\begin{table}[t]
\centering
\caption{Downstream policy success rates (\%) on LIBERO\_10 when training $\pi$0.5 with 500 real demonstrations plus generated data from different sources under a matched budget (1,500 videos).}
\label{tab:policy_learning}
\resizebox{0.62\textwidth}{!}{
\begin{tabular}{ccc}
\toprule
Real Data & Augmented Data & Success Rate (\%) \\
\midrule
500 & 0 (baseline) & 58.4 \\
500 & 1500 short (CogVideoX) & 44.6 \\
500 & 1500 long (CausVid) & 62.0 \\
500 & 1500 long (Genie-Envisioner) & 60.2 \\
500 & 1500 long (MAGI-1) & 57.6 \\
\midrule
500 & 500 long (LongScape)  & 62.8 \\
500 & 1000 long (LongScape) & 66.0 \\
500 & 1500 long (LongScape) & \textbf{68.2} \\
\bottomrule
\end{tabular}}
\end{table}

To verify that LongScape's long-horizon generation translates to practical embodied learning benefits, we evaluate its synthetic data in downstream policy training.
Using $\pi$0.5 on LIBERO\_10, we augment a baseline set of 500 real demonstrations with LongScape-generated trajectories of varying sizes.
We use the inverse dynamics model (IDM) from WoW~\citep{chi2025wow} to infer action sequences from the generated videos for downstream $\pi$0.5 policy training.
To isolate whether the benefit stems from horizon length or from the quality of LongScape's generation, we further compare against equal-budget augmentation with long-horizon videos generated by CausVid, Genie-Envisioner, and MAGI-1, as well as an equal volume of short clips (approx.\ 5\,s) from CogVideoX, representing typical short-horizon generation.

As shown in Table~\ref{tab:policy_learning}, LongScape-generated long videos consistently improve policy success rate, reaching a gain of 9.8 percentage points over the real-data-only baseline.
Two observations highlight the utility of LongScape-generated data.
First, LongScape augmentation raises policy success to 68.2\%, compared with 44.6\% for short CogVideoX clips, demonstrating the greater utility of LongScape-generated long-horizon data in this setting.
Second, horizon length alone does not account for the downstream gains: among the long-horizon generators, LongScape outperforms the best alternative (CausVid, 62.0\%) by 6.2 percentage points under the same data budget.
These results confirm that LongScape's stable long-horizon generation directly benefits embodied policy learning, rather than serving merely as a video quality benchmark.

%% file: sec/5_conclusion.tex
\section{Conclusion and Future Works}

In this work, we introduce LongScape, a novel framework for long-horizon embodied world modeling that adaptively integrates diffusion denoising with autoregressive generation. Our method leverages action priors to guide video chunk partitioning and employs a dynamic router to activate specialized generation experts. This approach enables the production of video tailored to the semantic context of each chunk, significantly enhancing visual quality and coherence. Experimental results demonstrate that LongScape outperforms existing approaches by maintaining stable and coherent video generation over extended horizons.

Despite these advancements, several directions merit further exploration. First, while the current action-guided chunking mechanism relies on heuristic rules, future work could investigate learning-based chunking strategies to achieve more flexible and task-aware video partitioning. Second, extending LongScape to support multi-view camera inputs would further enhance its applicability in complex real-world scenarios.

%% file: sec/6_appendix.tex
\renewcommand{\thesection}{}%
\renewcommand{\thesubsection}{\arabic{subsection}}%
\setcounter{subsection}{0}%
\section*{Appendix}

\subsection{Reproducibility statement} \label{sec:a1}
To ensure reproducibility, we have made our code and visualization cases publicly available at: \url{https://github.com/tsinghua-fib-lab/Longscape}.

\subsection{Details of Datasets} \label{appendix:dataset}
We conduct experiments on LIBERO and AGIBOT-World datasets, both focusing on long-horizon composite embodied manipulation tasks.
\begin{itemize}[leftmargin=*]
    \item  \textbf{LIBERO}~\citep{liu2023libero}: A benchmark for lifelong robot learning, including 130 diverse manipulation tasks organized into specialized suites (LIBERO-Spatial, LIBERO-Object, LIBERO-Goal, and LIBERO-100) to study knowledge transfer across variations in object spatial relationships, object types, and task goals. In our experiments, we performed training and evaluation on each of the 130 tasks individually.
    \item \textbf{AGIBOT-World}~\citep{bu2025agibot}: A large-scale real-world dataset collected with a wheeled dual-arm robot. It covers over 100 real-world scenarios across five core environments (household, catering, industrial, commercial, office) and contains a high proportion of long-horizon tasks. Our experiments utilize a subset focusing on two representative long-horizon task categories: folding clothes and pick-and-place operations.
\end{itemize}

\subsection{Details of Baselines} \label{appendix:baseline}
We compare our approach with seven advanced video generation world models, covering purely diffusion-based, autoregressive, and combined diffusion-autoregressive architectures, which are detailed as follows.
\begin{itemize}[leftmargin=*]
    \item \textbf{CogVideoX}~\citep{yang2024cogvideox}: An advanced open-source diffusion-based text-to-video model. Its core innovation is a DiT-based expert transformer that employs 3D full attention to effectively model complex spatiotemporal relationships, achieving high visual fidelity. We use the 5B parameter version for a fair comparison.
    \item \textbf{Genie}~\citep{bruce2024genie}: A large-scale autoregressive foundation world model. It operates by encoding video frames into discrete tokens and employs a spatial-temporal transformer for next-token prediction. As a seminal work, it demonstrates the potential of scaling autoregressive paradigms for diverse video generation tasks.
    \item \textbf{Cosmos-Predict2.5}~\citep{nvidia2025worldsimulationvideofoundation}: A DiT-based video world model trained with rectified flow. It unifies text-, image-, and video-conditioned generation to simulate and predict future world states, serving as a flow-based world model baseline for embodied video generation.
    \item \textbf{NOVA}~\citep{deng2024autoregressive}: A hybrid diffusion-autoregressive model designed for streaming video generation. It integrates per-frame diffusion denoising within an autoregressive framework, ensuring both high intra-frame detail and robust inter-frame temporal coherence across long sequences.
    \item \textbf{MAGI-1}~\citep{teng2025magi}: A state-of-the-art hybrid model for long video generation. It performs fixed-length diffusion denoising within an autoregressive chain, effectively balancing the visual quality advantages of diffusion models with the long-range consistency of autoregressive approaches.
    \item \textbf{CausVid}~\citep{zhuang2025video}: A video generation model that emphasizes causal consistency. It employs a teacher-student distillation framework, which transfers the knowledge from a pre-trained diffusion model to a fast autoregressive generator for efficient streaming generation.
    \item \textbf{Genie-Envisioner}~\citep{liao2025genie}: A unified world foundation platform specifically designed for embodied scenarios. It facilitates realistic video prediction for robotic manipulation by supporting multi-view camera inputs and action conditioning, serving as a strong baseline in the domain of embodied AI.
    
\end{itemize}

\subsection{Details of Metrics} \label{appendix:metrics}
We evaluate the generated videos using the following metrics:
\begin{itemize}[leftmargin=*]
\item \textbf{PSNR}: It evaluates the pixel-level similarity between generated and ground-truth frames.
\item \textbf{LPIPS}: It measures the image feature similarity of generated and ground truth frames.
\item \textbf{SSIM}: It assesses the brightness, contrast, and structural consistency between generated and ground truth frames.
\item \textbf{FVD}: It measures the distance between real and generated video feature distributions.
\item \textbf{Subject Consistency}: It measures the perceptual stability of the main subject across the video frames, following VBench~\citep{huang2024vbench} protocol.
\item \textbf{Background Consistency}: It assesses the temporal coherence and invariance of the background scene throughout the video, following VBench~\citep{huang2024vbench} protocol.
\end{itemize}

\subsection{Details of Implementation}
\label{appendix:implementation}

For both datasets, each DiT expert contains 5.57B parameters, initialized from the CogVideoX-5B checkpoint~\citep{yang2024cogvideox}. The dynamic router has 108.57M parameters for AGIBOT-World and 71.35M parameters for LIBERO.
We sample videos at 10\,Hz from LIBERO and AGIBOT-World, yielding approximately 30,000 training clips from each source.
Each expert is fine-tuned independently on its designated chunk type for 2 epochs using 4 NVIDIA H20 GPUs, requiring around 24 hours per expert.
The router is trained for a similar duration (24 hours) on 4 NVIDIA H20 GPUs.
Inference throughput measurements are conducted on a single NVIDIA GeForce RTX 4090D GPU, using CPU offloading to keep only the selected expert on the GPU.

\begin{algorithm}[h]
\caption{Algorithm of chunk partition}
\label{alg:chunk}
\textbf{Input:} Video frames $F$, action sequence $A$ for each frame \\
\textbf{Output:} Partitioned chunks set $S= \{S_1, S_2, \dots, S_k\}$
\begin{algorithmic}[1]
\State Divide video into base chunks $ C$ of 8 frames each 
\State Calculate $R_d = \max_f A_d(f)-\min_f A_d(f)$ and $\theta_d = \alpha R_d$ with $\alpha=0.5$ for each action dimension $d$
\State Isolate each base chunk whose gripper state changes as an independent chunk
\While {not all base chunks processed}:
\State Check candidate chunk $C_{1:n}$ of length n (1-4 base chunks):
\quad\If{$\exists d:\ \Delta A_{d,1:n} > \theta_d$}: add $C_{1:n}$ to $S$
\quad\ElsIf{n=4}: add $C_{1:n}$ to $S$
\quad\ElsIf{$C_{1:n}$ reaches the segment end}: add $C_{1:n}$ to $S$
\quad\Else: increase n and continue checking
\EndIf
\EndWhile
\end{algorithmic}
\end{algorithm}

\begin{figure*}[t]
\begin{center}
\includegraphics[width=\textwidth]{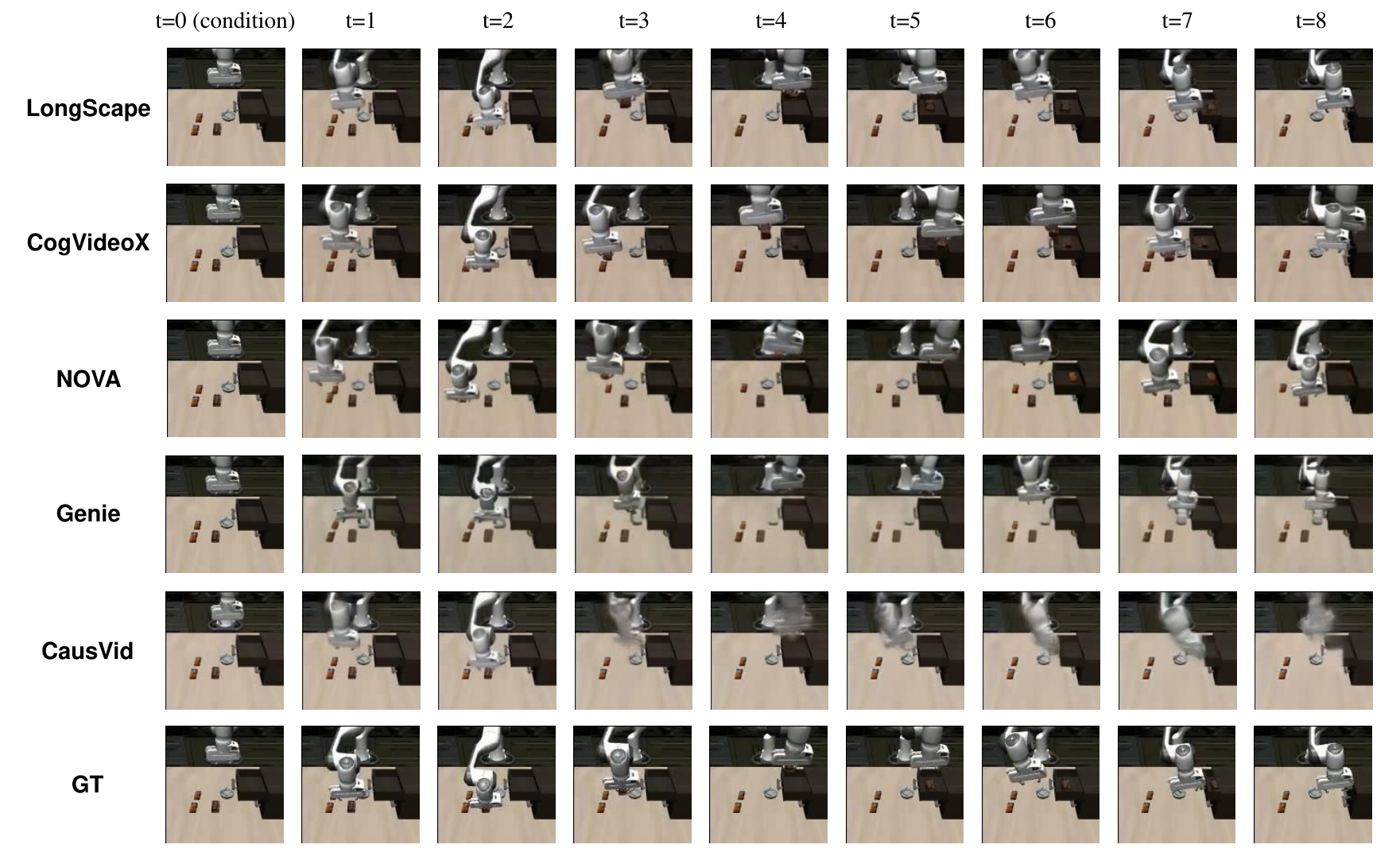}

\end{center}
\caption{Qualitative comparison on LIBERO dataset with text instruction ``put the chocolate pudding in the top drawer of the cabinet and close it''. Here we select several models with good results and sample eight frames from the generated long video sequences for better presentation.}
\label{fig:main_libero}
\end{figure*}

\begin{figure}[h]
\begin{center}
\includegraphics[width=\textwidth]{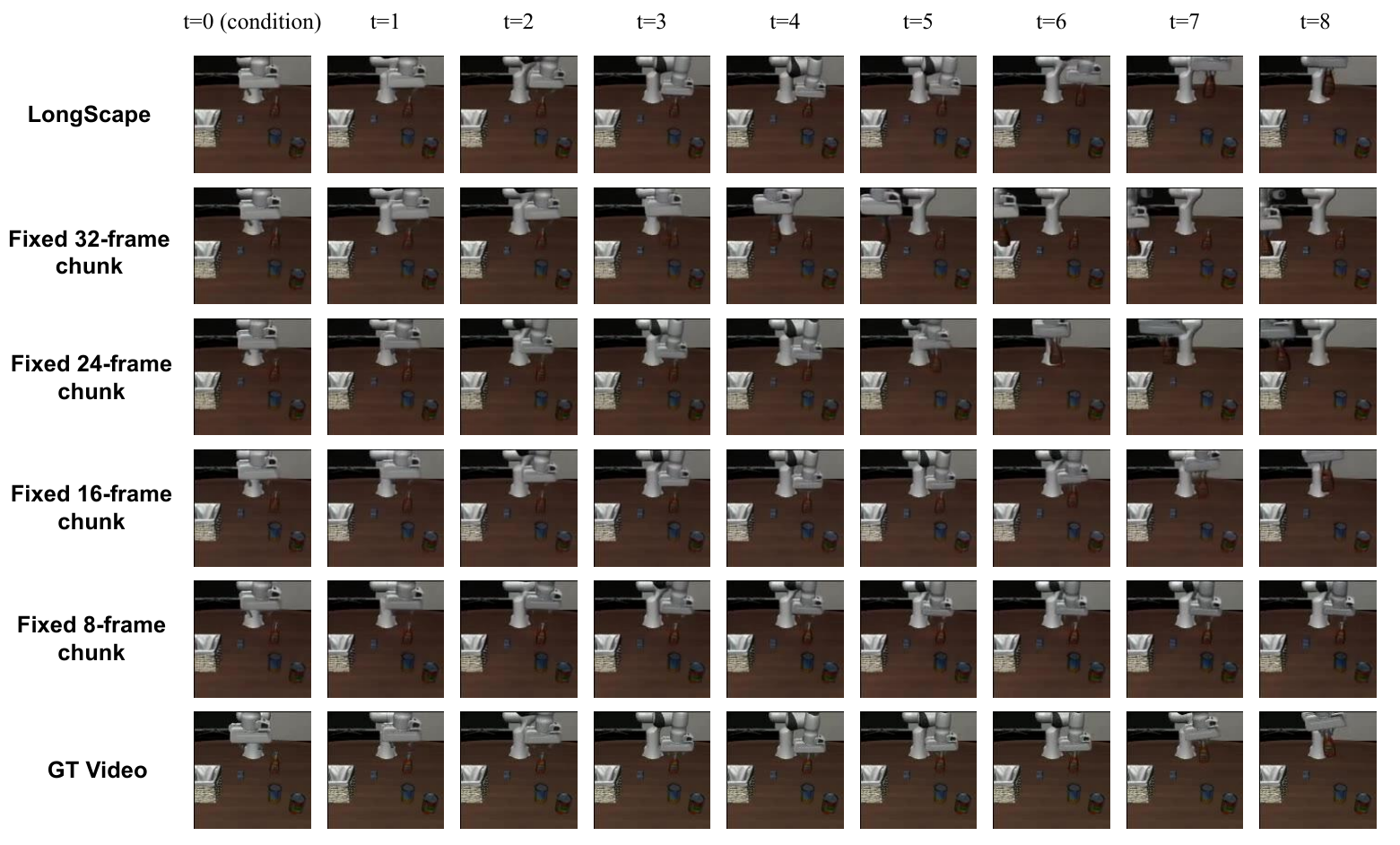}

\end{center}
\caption{Visualization of the ablation study on LIBERO dataset. We compare LongScape with variants using only fixed-length chunks. These fixed-length approaches often exhibit discontinuous or physically implausible actions. In contrast, LongScape, by incorporating an adaptive expert mechanism, ensures action coherence and stable generation.}
\label{fig:ab_libero}
\end{figure}

\subsection{Sensitivity Analysis of the Motion Threshold Ratio $\alpha$} \label{sec:threshold}
We vary the threshold ratio $\alpha \in \{0.3, 0.5, 0.7\}$ in $\theta_d = \alpha R_d$ on the complete LongScape pipeline. All three settings outperform the best single fixed-length expert, with $\alpha = 0.5$ yielding the strongest overall performance.
\begin{table}[h]
\centering
\caption{Performance across different threshold ratios $\alpha$ on LIBERO.}
\resizebox{0.6\textwidth}{!}{
\begin{tabular}{ccccc}
\toprule
$\alpha$ & PSNR$\uparrow$ & LPIPS$\downarrow$ & SSIM$\uparrow$ & FVD$\downarrow$ \\
\midrule
0.3 & 19.865 & 0.1246 & 0.7883 & 158.93 \\
0.5 & \textbf{19.977} & \textbf{0.1231} & \textbf{0.7883} & \textbf{153.72} \\
0.7 & 19.923 & 0.1243 & 0.7856 & 172.74 \\
\hline
Best single DiT & 19.755 & 0.1285 & 0.7845 & 180.31 \\
\bottomrule
\end{tabular}}
\end{table}

\subsection{Long-Horizon Stability Analysis} \label{sec:longhorizon}
The main benchmarks evaluate rollouts of approximately 20 seconds. To evaluate long-horizon generation, we generate 1,000-frame rollouts on AGIBOT-World and compute the six evaluation metrics on prefixes of 200, 400, 600, 800, and 1,000 frames. Table~\ref{tab:longhorizon} reports the mean results across the test rollouts.

\begin{table}[h]
\centering
\caption{Average generation quality across AGIBOT-World test rollouts, evaluated on prefixes ranging from 200 to 1,000 frames.}
\label{tab:longhorizon}
\resizebox{0.75\textwidth}{!}{
\begin{tabular}{ccccccc}
\toprule
Frames $T$ & PSNR$\uparrow$ & LPIPS$\downarrow$ & SSIM$\uparrow$ & FVD$\downarrow$ & \makecell{Subject\\ Consistency$\uparrow$} & \makecell{Background \\Consistency$\uparrow$} \\
\midrule
200 & 16.70 & 0.315 & 0.700 & 191.0 & 0.935 & 0.936 \\
400 & 16.19 & 0.351 & 0.689 & 222.5 & 0.924 & 0.931 \\
600 & 15.75 & 0.380 & 0.664 & 271.9 & 0.915 & 0.926 \\
800 & 15.71 & 0.389 & 0.655 & 269.5 & 0.914 & 0.920 \\
1000 & 15.65 & 0.390 & 0.648 & 273.9 & 0.910 & 0.916 \\
\bottomrule
\end{tabular}}
\end{table}

As shown in Table~\ref{tab:longhorizon}, LongScape sustains high generation quality and temporal consistency over 1,000-frame rollouts. At 1,000 frames, the average PSNR reaches 15.65\,dB, with Subject and Background Consistency scores of 0.910 and 0.916, respectively. These results demonstrate LongScape's ability to maintain coherent long-horizon generation across the evaluated test rollouts.

\subsection{Ablation of Inference-Time Length Selection} \label{sec:length_policy}
To evaluate inference-time length selection, we keep the four trained DiT experts fixed and vary only the rule that selects the next chunk length and expert on LIBERO. We compare the action-supervised router against uniformly random lengths from $\{8,16,24,32\}$, a non-contextual round-robin schedule, and a visual-change heuristic that selects short chunks when large frame differences are detected.

\begin{table}[t]
\centering
\caption{Ablation of inference-time chunk length policies on LIBERO, with the four trained DiT experts kept fixed. Only the rule that selects the next chunk length (and thus the expert) is varied.}
\label{tab:length_policy}
\resizebox{0.83\textwidth}{!}{
\begin{tabular}{ccccccc}
\toprule
\textbf{Length Policy} & PSNR$\uparrow$ & LPIPS$\downarrow$ & SSIM$\uparrow$ & FVD$\downarrow$ & \makecell{Subject\\ Consistency$\uparrow$} & \makecell{Background \\Consistency$\uparrow$} \\
\midrule
\textbf{Router (LongScape, action-supervised)} & \textbf{19.977} & \textbf{0.1231} & \textbf{0.7883} & \textbf{153.72} & \textbf{0.9297} & \textbf{0.9622} \\
Random length $\in\{8,16,24,32\}$ & 19.538 & 0.1329 & 0.7792 & 176.38 & 0.9100 & 0.9612 \\
Round-robin (non-contextual) & 19.538 & 0.1347 & 0.7780 & 183.24 & 0.9076 & 0.9596 \\
Visual-change heuristic & 19.602 & 0.1345 & 0.7797 & 189.59 & 0.9088 & 0.9596 \\
\bottomrule
\end{tabular}}
\end{table}

As shown in Table~\ref{tab:length_policy}, the learned router outperforms all three alternatives across all six metrics. With the same expert bank, contextual routing achieves an FVD of 153.72 and Subject Consistency of 0.9297, compared with FVD values of 176.38--189.59 and Subject Consistency scores of 0.9076--0.9100 for the alternative policies. These results demonstrate the benefit of learned contextual routing for inference-time length selection.

\subsection{Router Modality Ablation} \label{sec:router_modality}
To assess whether the text instruction assists expert selection, we ablate the router input modalities on the LIBERO validation protocol, keeping all trained weights fixed:

\begin{table}[h]
\centering
\caption{Router accuracy under different input modalities on LIBERO. The full-router reference accuracy is taken from the original evaluation reported in the main text.}
\label{tab:router_modality}
\resizebox{0.5\textwidth}{!}{
\begin{tabular}{cc}
\toprule
Router Input & Accuracy (\%) \\
\midrule
Text + Vision (full) & \textbf{91.4} \\
Vision-only & 24.97 \\
Text-only & 12.76 \\
\bottomrule
\end{tabular}}
\end{table}

As shown in Table~\ref{tab:router_modality}, removing either modality causes a severe drop in expert-selection accuracy. Vision-only falls to near chance for a four-way decision, while text-only performs even worse. The poor text-only performance is expected: textual instructions such as ``put both the alphabet soup and the tomato sauce in the basket'' are distributed nearly uniformly across all four chunk lengths and offer little discriminative temporal information, whereas visual grounding is essential for localizing the current phase of execution. The router therefore relies on the learned fusion of the textual instruction and the current visual context.